\documentclass{article}

\PassOptionsToPackage{numbers,sort&compress}{natbib}
\usepackage[dblblindworkshop,final]{neurips_2026}
\workshoptitle{Machine Learning and the Physical Sciences (ML4PS 2026)}

\usepackage[utf8]{inputenc}
\usepackage[T1]{fontenc}
\usepackage{amsmath,amssymb}
\usepackage{amsthm}
\usepackage{bm}
\usepackage{algorithm}
\usepackage{algpseudocode}
\usepackage{booktabs}
\usepackage{flafter}
\usepackage{float}
\usepackage{graphicx}
\usepackage{multirow}
\usepackage{nicefrac}
\usepackage{subcaption}
\usepackage{microtype}
\usepackage{mathtools}
\usepackage{placeins}
\usepackage{xcolor}
\usepackage{url}
\usepackage{hyperref}

\usepackage{tikz}
\usepackage{xspace}
\usetikzlibrary{arrows.meta,backgrounds,calc,fit,positioning}

\graphicspath{{figures/generated/}{figures/}}
\DeclareMathOperator{\Log}{Log}
\DeclareMathOperator{\Exp}{Exp}
\newcommand{\HH}{\mathcal{H}}
\newcommand{\Proj}{\Pi}
\newcommand{\dist}{d_g}
\newcommand{\R}{\mathbb{R}}

\newcommand{\jetnet}{{\textsc{JetNet}}\xspace}

\definecolor{mlpfill}{HTML}{B7CAE8}
\definecolor{mlpdraw}{HTML}{2459A6}
\definecolor{sumfill}{HTML}{FFE2CC}
\definecolor{sumdraw}{HTML}{D95F02}
\definecolor{geomfill}{HTML}{C9C7C7}
\definecolor{geomdraw}{HTML}{4D4D4D}
\definecolor{groupdraw}{HTML}{9B9B9B}

\newtheorem{theorem}{Theorem}

\title{\textsc{JetFUEL}: A physics-informed recipe for particle cloud generation with flow matching}

\author{%
  \textbf{Arjun Sharma\textsuperscript{1},
  Zichun Hao\textsuperscript{1},
  Raghav Kansal\textsuperscript{2}} \\
  \textbf{Abhijith Gandrakota\textsuperscript{3},
  Jennifer Ngadiuba\textsuperscript{3},
  Maria Spiropulu\textsuperscript{1}} \\[0.8em]
  \textsuperscript{1}Division of Physics, Mathematics and Astronomy \\
  California Institute of Technology, Pasadena, CA 91125 \\[0.4em]
  \textsuperscript{2}Genesis Molecular AI \\[0.4em]
  \textsuperscript{3}Particle Physics Division \\
  Fermi National Accelerator Laboratory, Batavia, IL 60510
}
\begin{document}

\maketitle

\begin{abstract}
    Deep generative models are increasingly used across scientific domains to model natural data distributions efficiently. In high-energy physics, previous work has successfully trained generative models of jets: collimated sprays of particles produced in collisions such as those at the CERN Large Hadron Collider. We present \textsc{JetFUEL} (Jet Flow matching with eqUivariancE under the Lorentz group), a generative modeling pipeline for jets with rich physical priors. First, its learned velocity field is Lorentz-equivariant and uses only scalar and four-vector features. Second, the model is trained with Riemannian flow matching, which constrains paths, targets, and sampling trajectories to a regulated mass-shell hyperboloid in momentum space, thereby preventing unphysical states during generation. We find that, in this physics-informed setting, the training recipe is sensitive to the prior distribution, access to jet-axis information, and the coupling between prior and data particle clouds. We train a single model across multiple jet types and use classifier-free guidance to generate jets conditionally at inference time; to our knowledge, this is the first application of classifier-free guidance to jet generation. \textsc{JetFUEL} is competitive with prior work on jet generation: on \jetnet, a single class-conditional model with $\sim 0.6$M parameters reaches high fidelity on the standard evaluation metrics for gluon, light quark, and top quark jets. We release our code here: \href{https://github.com/ArjunS07/jetfuel}{github.com/ArjunS07/jetfuel}.
\end{abstract}

\section{Introduction}
\label{sec:introduction}

High-energy physics (HEP) research leverages simulations of billions of events\footnote{An event in HEP is a set of observed particles with well-defined momenta, but not generally well-defined positions.} to inform experiments and test theoretical predictions~\cite{apostolakis2022detector,karagiorgi2022machine}. In recent years, there has been increasing interest in using deep generative models to simulate events in an end-to-end fashion~\cite{guest2018deep,kansalParticleCloudGeneration2022,mikuniFastPointCloud2023,buhmannEPiCGANEquivariantPoint2023,oliveiraLearningParticlePhysics2017,butterMachineLearningLHC2023,FastSimulationATLAS2020,touranakouParticlebasedFastJet2022,leighPCJeDiDiffusionParticle2024,kachAttentionMeanFieldsParticle2023a,kachJetFlowGeneratingJets2022}. These deep generative models can accelerate simulation by many orders of magnitude~\cite{kansalParticleCloudGeneration2022} compared to traditional techniques, which provide comprehensive simulations but frequently remain approximations to nature. In this work, we study deep generative models in the context of \textit{jets}: collimated high-energy sprays of particles produced at the CERN Large Hadron Collider (LHC)~\cite{oliveiraLearningParticlePhysics2017,kansal2023jetnet}.

Deep generative modeling in scientific domains raises design considerations not typically encountered in the original domains for which these techniques were developed, such as language and vision. A popular approach~\cite{bronstein2021geometric, jumper2021highly} is to explicitly incorporate scientific inductive biases into model architectures, thus restricting the learnable function space to potentially improve physical plausibility and interpretability by drawing on domain knowledge. In the context of HEP, we know that jets are governed by the laws of special relativity, so their constituents' four-momenta transform in a well-defined way under Lorentz boosts and rotations~\cite{liDoesLorentzsymmetricDesign2024}, which together form the Lorentz group.
Additionally, jet constituents carry no intrinsic ordering and models should thus respect permutation symmetry~\cite{kansal2023jetnet, larkoski2020jet}. 
Machine learning methods for jets have widely adopted inductive biases by utilizing permutation-invariant architectures~\cite{kansalParticleCloudGeneration2022,buhmannEPiCGANEquivariantPoint2023,leighPCJeDiDiffusionParticle2024} and, more recently, architectures that are equivariant under the Lorentz group~\cite{gongEfficientLorentzEquivariant2022,haoLorentzGroupEquivariant2023,spinner2024lorentz,spinner2026lorentz}. Extending this line of work, we develop a simple Lorentz-equivariant architecture with $\sim 0.6$M parameters for generative modeling of jets. We integrate this with further physical knowledge by restricting generative modeling to a physics-informed manifold using the Riemannian Flow Matching (RFM) framework~\cite{chen2024flow}. Ours is the first generative model for jets that simultaneously respects Lorentz symmetries as well as the jet data manifold throughout the training and generation process.

We introduce an end-to-end recipe for jet generation that builds physical knowledge into the architecture, the training objective, and inference. Our velocity field adapts the Lorentz-equivariant and permutation-invariant LorentzNet architecture~\cite{gongEfficientLorentzEquivariant2022}.
Unlike previous Lorentz-equivariant generative models, our model uses only Lorentz scalars and four-vectors, rather than geometric-algebra multivectors~\cite{spinner2024lorentz} or local frame canonicalization~\cite{spinner2026lorentz}. We additionally enforce the known on-shell relationship between energy and momentum by running RFM on the mass shell, such that every intermediate and final constituent has positive energy by construction. 
Because the architecture is rotation-equivariant, we find it critical to supply the jet direction explicitly through the choice of prior and by passing the jet four-vector to the network. We train a single model across diverse jet types, utilizing classifier-free guidance to control global jet characteristics at inference time, and achieve competitive results with MPGAN~\cite{kansalParticleCloudGeneration2022} on gluon, light quark, and top quark jets while respecting physical symmetries and constraints.


\section{Method}
\label{sec:method}

\subsection{Background}

\textbf{Riemannian flow matching}. Flow matching \cite{lipmanFlowMatchingGenerative2023} is a generative modeling technique that transports a tractable source distribution to a target data distribution. 
Riemannian Flow Matching (RFM) \cite{chen2024flow} extends this framework to a general Riemannian manifold $(\mathcal{M},g)$, where $g$ is a positive-definite metric, by constraining the vector field to the manifold's tangent bundle. Given a prior $p_0$ on $\mathcal{M}$ and data samples $x_1 \sim q$, RFM learns a vector field $V_\theta(t,x)\in T_x\mathcal{M}$ whose flow transports $p_0$ to
$p_1\approx q$. For $t\sim\mathcal{U}[0,1]$ and $x_t\sim p_t(\cdot\mid x_1)$, the objective is
\begin{equation}
    \mathcal{L}_{\mathrm{RCFM}}(\theta)
    =
    \mathbb{E}_{t,\,q(x_1),\,p_t(x_t\mid x_1)}
    \left[
        \left\|
            V_\theta(t,x_t)-u_t(x_t\mid x_1)
        \right\|_{g}^{2}
    \right]
    \label{eq:rfm-loss}
\end{equation}
where $u_t(x_t\mid x_1)\in T_{x_t}\mathcal{M}$ is the target velocity of the conditional probability path and $\|\cdot\|_g$ is the norm induced by $g$.  

In this work, we focus on cases in which we can use geodesic conditional paths. In this case, the map $\Log_x(y)\in T_x\mathcal M$ gives the initial tangent vector of the geodesic from $x$ to $y$, and the map $\Exp_x:T_x\mathcal M\rightarrow\mathcal M$ maps a tangent vector at $x$ back onto the manifold. Now, using the interpolation schedule $\kappa(t)=1-t$, the conditional path and its velocity~\cite{chen2024flow} for flow matching at training time are
\begin{align}
    x_t &= \Exp_{x_1}\!\left((1-t)\Log_{x_1}(x_0)\right)\\
    u_t(x_t\mid x_1) &= \frac{\Log_{x_t}(x_1)}{1-t}.
\end{align}
 
At inference time, we integrate the learned vector field using exponential Euler updates, which ensure that each updated state remains on $\mathcal M$: $x_{k+1}=\Exp_{x_k}\!\left(\Delta t\,V_\theta(t_k,x_k)\right)$.

\textbf{Jets}. Following previous work~\cite{kansal2023jetnet}, we represent jets as particle clouds with variable multiplicity, with each particle defined by its four-momentum $(E,p_x,p_y,p_z) = (p^0, p^1, p^2, p^3)$. We define the Minkowski inner product of two four-momentum vectors $p$ and $q$ as 
\begin{equation}
    \langle p, q \rangle_M = p^0 q^0 - \sum\limits_{a=1}^3 p^a q^a.
\end{equation}
The Minkowski norm $\lVert p \rVert_M^2 = \langle p, p \rangle_M$ of a four-vector defines its invariant mass squared, which is preserved under transformations of the Lorentz group. 

Jet constituents are considered massless, i.e., $\lVert p \rVert_M = 0$, and their energies $E$ are always positive. This restricts each constituent four-momentum to the ``forward light cone'' $C^+=\left\{p\in\mathbb R^{1,3}:\langle p,p\rangle_M=0,\ E>0\right\}$ and means that constituents are fully determined by their three-momentum, since $\lVert p \rVert_M = 0 \implies E^2 = p_x^2 + p_y^2 + p_z^2$.

\subsection{Constraining flow matching for jet generation with RFM}

\textbf{Defining a hyperboloid for RFM on jets}. We use our knowledge of these physical constraints on jet constituents to inform jet generation through RFM. Standard RFM cannot be applied directly to $C^+$ using the induced Minkowski metric as $g$ since this metric is not positive definite on the tangent space of $C^+$.\footnote{A short proof is provided in \autoref{appendix:rfm}: for any $p\in C^+$ and $\lambda>0$ with $\lambda\neq1$, the nonzero displacement $(\lambda-1)p$ is tangent to the light cone but has zero Minkowski norm.}

We instead make RFM tractable by using the fact that a massless constituent $p$ is fully determined by its three-momentum. Writing \(\mathbf{p}=(p_x,p_y,p_z)\), we have $p = \left(\lVert\mathbf{p}\rVert_2,\mathbf{p} \right) \in C^+$. We lift it to a hyperboloid $
    \mathcal{H}_m=\{p\in\R^{1,3}:\langle p,p\rangle_M=m^2,\ p^0>0\}$ by preserving \(\mathbf{p}\) and changing only its energy with the following map: $\left(\lVert\mathbf{p}\rVert_2,\mathbf{p}
\right)\longmapsto\left(\sqrt{\lVert\mathbf{p}\rVert_2^2+m^2},\mathbf{p}\right)$. We can then perform RFM training and generation on \(\mathcal{H}_m\). This hyperboloid \(\mathcal{H}_m\) conveniently admits a positive-definite Riemannian metric derived from the ambient Minkowski metric, along with closed-form expressions for the geodesic distance, logarithmic map, and exponential map\footnote{We summarize these expressions and their derivation in Appendix~\ref{appendix:hyperboloid-geometry}.}. 

After integration, we map each constituent \(p_m=(E_m,\mathbf{p})\in\mathcal{H}_m\) back to a massless state using $\left(
E_m,\mathbf{p}
\right)
\longmapsto
\left(
\lVert\mathbf{p}\rVert_2,\mathbf{p}
\right)$.

Our method exploits the intrinsic low-dimensional structure of the data to facilitate learning and guarantees that $E = \lVert\mathbf{p}\rVert_2 > 0$ at all times. Since jets consist of clouds of four-vector constituents, we construct the path and target independently for each constituent and average the RFM loss over constituents.

\textbf{Using geodesics for particle-cloud coupling}. Following \cite{klein2023equivariant}, we facilitate training by exploiting the permutation invariance of particle clouds. At training time, we use the Hungarian algorithm to find a permutation that minimizes the total pairwise distance between the constituents of $x_0$ and $x_1$.



\subsection{A simple Lorentz-equivariant architecture for generative modeling}

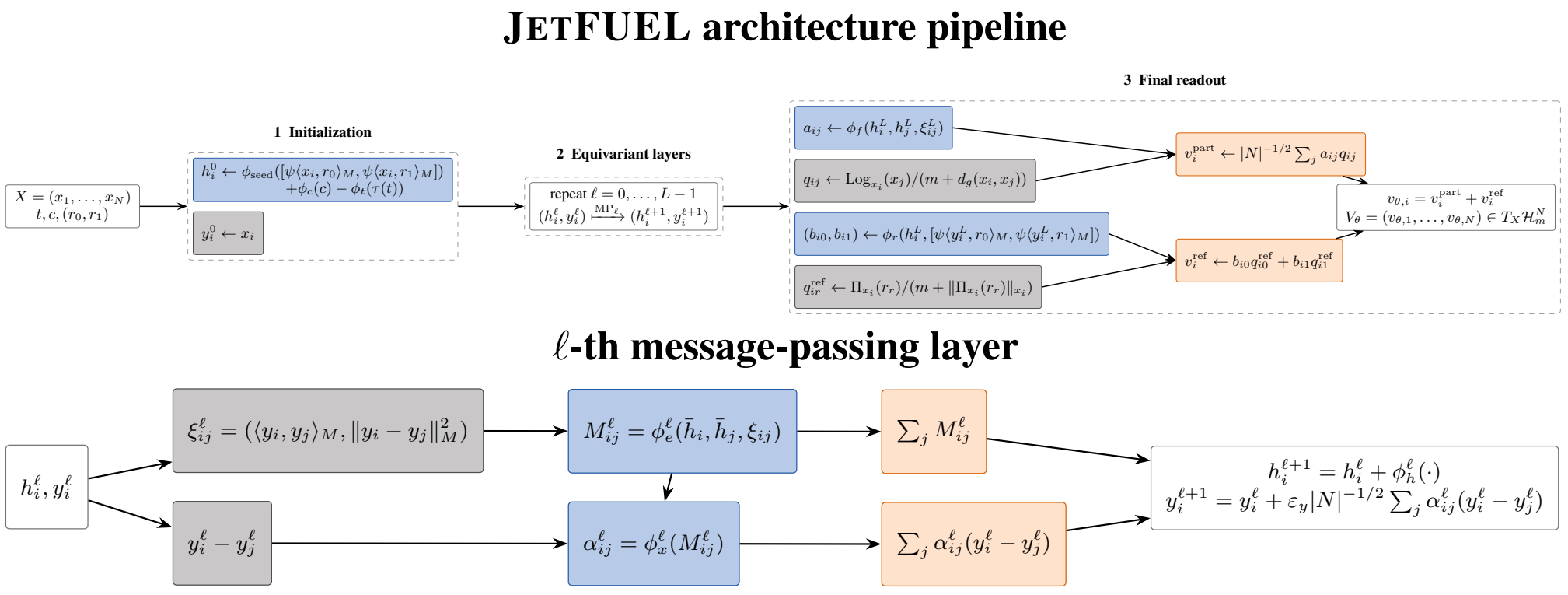
\begin{figure}[H]
  \centering
  \begin{subfigure}{\textwidth}
    \centering
    {\bfseries \textsc{JetFUEL} architecture pipeline\par}
    \vspace{0.4em}
    \resizebox{0.95\linewidth}{!}{
\begin{tikzpicture}[
  font=\rmfamily\small,
  >={Stealth[length=2.2mm,width=1.6mm]},
  signal/.style={->,draw=black,line width=0.7pt},
  mlp/.style={draw=mlpdraw,fill=mlpfill,rounded corners=1.5pt,
    minimum height=9mm,align=center,inner xsep=5pt,inner ysep=3pt},
  sum/.style={draw=sumdraw,fill=sumfill,rounded corners=1.5pt,
    minimum height=9mm,align=center,inner xsep=5pt,inner ysep=3pt},
  geometry/.style={draw=geomdraw,fill=geomfill,rounded corners=1.5pt,
    minimum height=9mm,align=center,inner xsep=5pt,inner ysep=3pt},
  result/.style={draw=black!55,fill=white,rounded corners=1.5pt,
    minimum height=9mm,align=center,inner xsep=5pt,inner ysep=3pt},
  stage/.style={draw=groupdraw,dashed,rounded corners=2pt,inner sep=3pt},
  label/.style={font=\rmfamily\bfseries\small},
  rowgap/.style args={of #1}{below=2mm of #1,anchor=north west}
]

\node[mlp] (inith) at (0,0)
  {$h_i^0\gets\phi_{\rm seed}\!\left([\psi\langle x_i,r_0\rangle_M,
     \psi\langle x_i,r_1\rangle_M]\right)$\\
   $\mathrel{\phantom{h_i^0\gets{}}}+\phi_c(c)-\phi_t(\tau(t))$};
\node[geometry,rowgap=of inith.south west] (inity) {$y_i^0\gets x_i$};
\node[stage,fit=(inith)(inity)] (initgroup) {};
\node[label,above=2mm of initgroup] {1\enspace Initialization};

\node[result,left=10mm of initgroup] (input)
  {$X=(x_1,\ldots,x_N)$\\$t,c,(r_0,r_1)$};

\node[result,right=15mm of initgroup] (layers)
  {repeat $\ell=0,\ldots,L-1$\\
   $(h_i^\ell,y_i^\ell)\xmapsto{\mathrm{MP}_\ell}
   (h_i^{\ell+1},y_i^{\ell+1})$};
\node[stage,fit=(layers)] (layergroup) {};
\node[label,above=2mm of layergroup] {2\enspace Equivariant layers};

\node[mlp,right=15mm of layergroup,yshift=16.5mm] (partweights)
  {$a_{ij}\gets\phi_f(h_i^L,h_j^L,\xi_{ij}^L)$};
\node[geometry,rowgap=of partweights.south west] (partdirs)
  {$q_{ij}\gets\Log_{x_i}(x_j)/(m+\dist(x_i,x_j))$};
\node[mlp,rowgap=of partdirs.south west] (refweights)
  {$(b_{i0},b_{i1})\gets\phi_r(h_i^L,[\psi\langle y_i^L,r_0\rangle_M,
    \psi\langle y_i^L,r_1\rangle_M])$};
\node[geometry,rowgap=of refweights.south west] (refdirs)
  {$q_{ir}^{\rm ref}\gets\Proj_{x_i}(r_r)/(m+\|\Proj_{x_i}(r_r)\|_{x_i})$};

\coordinate (partmid) at ($(partweights)!0.5!(partdirs)$);
\coordinate (refmid) at ($(refweights)!0.5!(refdirs)$);
\coordinate (sumcolumn) at ($(refweights.east)+(14mm,0)$);
\node[sum,anchor=west] (partsum) at (sumcolumn |- partmid)
  {$v_i^{\rm part}\gets|N|^{-1/2}\sum_j a_{ij}q_{ij}$};
\node[sum,anchor=west] (refsum) at (sumcolumn |- refmid)
  {$v_i^{\rm ref}\gets b_{i0}q_{i0}^{\rm ref}+b_{i1}q_{i1}^{\rm ref}$};

\coordinate (combinemid) at ($(partsum)!0.5!(refsum)$);
\node[result,right=15mm of combinemid] (combine)
  {$v_{\theta,i}=v_i^{\rm part}+v_i^{\rm ref}$\\
   $V_\theta=(v_{\theta,1},\ldots,v_{\theta,N})\in T_X\mathcal{H}_m^N$};
\node[stage,fit=(partweights)(partdirs)(partsum)(refweights)(refdirs)(refsum)(combine)]
  (readoutgroup) {};
\node[label,above=2mm of readoutgroup] {3\enspace Final readout};

\draw[signal] (input.east) -- (initgroup.west |- input.east);
\draw[signal] (initgroup.east) -- (layergroup.west |- initgroup.east);
\draw[signal] (layergroup.east) -- (readoutgroup.west |- layergroup.east);

\draw[signal] (partweights.east) -- (partsum.west);
\draw[signal] (partdirs.east) -- (partsum.west);
\draw[signal] (refweights.east) -- (refsum.west);
\draw[signal] (refdirs.east) -- (refsum.west);
\draw[signal] (partsum) -- (combine);
\draw[signal] (refsum) -- (combine);

\end{tikzpicture}}
  \end{subfigure}

  \vspace{0.3em}
  \begin{subfigure}{\textwidth}
    \centering
    {\bfseries $\ell$-th message-passing layer\par}
    \vspace{0.4em}
    \resizebox{0.95\linewidth}{!}{
\begin{tikzpicture}[
  font=\rmfamily\small,
  >={Stealth[length=2.2mm,width=1.6mm]},
  signal/.style={->,draw=black,line width=0.7pt},
  mlp/.style={draw=mlpdraw,fill=mlpfill,rounded corners=1.5pt,
    minimum height=12mm,align=center,inner xsep=6pt},
  sum/.style={draw=sumdraw,fill=sumfill,rounded corners=1.5pt,
    minimum height=12mm,align=center,inner xsep=6pt},
  geometry/.style={draw=geomdraw,fill=geomfill,rounded corners=1.5pt,
    minimum height=12mm,align=center,inner xsep=6pt},
  result/.style={draw=black!55,fill=white,rounded corners=1.5pt,
    minimum height=12mm,align=center,inner xsep=6pt},
  rowgap/.style args={of #1}{below=4mm of #1,anchor=north west}
]

\node[geometry] (inv) at (0,0)
  {$\xi_{ij}^\ell=(\langle y_i,y_j\rangle_M,\|y_i-y_j\|_M^2)$};
\node[geometry,rowgap=of inv.south west] (disp) {$y_i^\ell-y_j^\ell$};
\node[fit=(inv)(disp),inner sep=0pt] (geometrycolumn) {};
\node[result,left=12mm of geometrycolumn] (state) {$h_i^\ell,y_i^\ell$};

\node[mlp,right=12mm of geometrycolumn,yshift=8mm] (edge)
  {$M_{ij}^\ell=\phi_e^\ell(\bar h_i,\bar h_j,\xi_{ij})$};
\node[mlp,rowgap=of edge.south west] (coeff)
  {$\alpha_{ij}^\ell=\phi_x^\ell(M_{ij}^\ell)$};
\node[fit=(edge)(coeff),inner sep=0pt] (learnedcolumn) {};

\node[sum,right=12mm of learnedcolumn,yshift=8mm] (sumh)
  {$\sum_jM_{ij}^\ell$};
\node[sum,rowgap=of sumh.south west] (sumy)
  {$\sum_j\alpha_{ij}^\ell(y_i^\ell-y_j^\ell)$};
\node[fit=(sumh)(sumy),inner sep=0pt] (sumcolumn) {};

\node[result,right=12mm of sumcolumn] (update)
  {$h_i^{\ell+1}=h_i^\ell+\phi_h^\ell(\cdot)$\\
   $y_i^{\ell+1}=y_i^\ell+\varepsilon_y|N|^{-1/2}
     \sum_j\alpha_{ij}^\ell(y_i^\ell-y_j^\ell)$};

\draw[signal] (state) -- (inv);
\draw[signal] (state) -- (disp);
\draw[signal] (inv) -- (edge);
\draw[signal] (edge) -- (sumh);
\draw[signal] (edge) -- (coeff);
\draw[signal] (disp) -- (coeff);
\draw[signal] (coeff) -- (sumy);
\draw[signal] (sumh) -- (update);
\draw[signal] (sumy) -- (update);

\end{tikzpicture}}
  \end{subfigure}
  \caption{The \textsc{JetFUEL} architecture pipeline (top) and its
    $\ell$-th message-passing layer (bottom), modified from the architecture of \cite{gongEfficientLorentzEquivariant2022}. Following \cite{gongEfficientLorentzEquivariant2022}, the model updates Lorentz vectors $y$ only as linear combinations of other vectors, with coefficients that depend on arbitrary functions of Lorentz-scalar quantities. Here, $\phi$ always represents a learnable multilayer perceptron, and $\Pi_{x_i}$ is the tangent projection operator defined in Appendix~\ref{appendix:tangent-projection}, and $\psi$ is a signed-log transform defined in Appendix~\ref{appdx:algorithm-description}.
}
\label{fig:jetfuel-architecture}
\end{figure}

We learn a time-dependent field parametrized by a model $V_\theta(x_t, t, N, c, r_0, r_1)$, where $N$ corresponds to the jet multiplicity, $c$ is a vector containing the jet type, total transverse momentum $p_{T,J}$, and total mass $m_J$; $r_0 = (1, 0, 0, 0)$; and $r_1 = p_J$, the jet axis. We implement this model by extending the permutation-invariant and Lorentz-equivariant LorentzNet architecture~\cite{gongEfficientLorentzEquivariant2022} to generative modeling.

Our architecture is shown in \autoref{fig:jetfuel-architecture}. Its primary contributions over previous literature \cite{gongEfficientLorentzEquivariant2022} are (1)~explicitly accounting for the global jet direction without violating Lorentz equivariance; (2)~seeding latent scalar states to facilitate message passing, even though the final output does not require them; and (3)~showing that it is possible for a Lorentz-equivariant architecture to be sufficiently expressive for generative modeling without utilizing machinery such as geometric algebra multivectors or local frame canonicalization \cite{spinner2024lorentz,spinner2026lorentz}. Additionally, our architecture is designed to complement RFM by using the on-manifold intermediate inputs as directions for the final output without restricting the internal vector state during message passing. We provide a complete model specification in \autoref{alg:network} in \autoref{appdx:network-architecture}.

\subsection{Supplying jet direction}
\label{sec:jet-direction}
A rotation-equivariant flow initialized from a rotation-invariant prior can generate only a rotation-invariant distribution (see \autoref{thm:equivariant-flow-preserves-invariance} in Appendix~\ref{appdx:equivariant-flow-invariance} for a short proof). However, the conditional distribution of jet constituents is not invariant under arbitrary rotations when the jet axis is held fixed~\cite{murnane2026equivariant,spinner2024lorentz}. We therefore pass the jet-axis four-momentum $r_1$ to the model (\autoref{fig:jetfuel-architecture}). We also use an axis-aligned conditional prior: given the jet-axis pseudorapidity $\eta_J$ and azimuth $\phi_J$, we sample particle positions around the jet axis as $\eta_i=\eta_J+\Delta\eta_i$ and $\phi_i=\phi_J+\Delta\phi_i$, with $\Delta\eta_i,\Delta\phi_i\stackrel{\rm iid}{\sim}\mathcal{N}(0,0.15^2)$. The other components of the prior are sampled as described in Appendix~\ref{sec:physical-prior}.

\subsection{Classifier-free guidance}

We train one model to generate diverse jet types and structures by utilizing classifier-free guidance (CFG) \cite{hoClassifierFreeDiffusionGuidance2022}. In practice, CFG is implemented at training time by randomly dropping out, with some probability $b$, the conditioning vector $c$ supplied to the network for a particular jet and replacing it with a null conditioning vector $\varnothing$. In our experiments, we follow previous literature~\cite{peebles2023scalable} and make the vector $\varnothing$ a learnable parameter, obviating the need for a hand-designed null vector which may spuriously clash with physically meaningful values. 

At inference time, CFG strengthens conditional generation by calculating a guided velocity field as \begin{equation}
\widetilde V_\theta(x_t,t,N,c,r_0,r_1)
=
V_\theta(x_t,t,N,c,r_0,r_1)
+
w\left[
V_\theta(x_t,t,N,c,r_0,r_1)
-
V_\theta(x_t,t,N,\varnothing,r_0,r_1)
\right],
\end{equation}
where $w$ is a weight amplifying the strength of the conditioning-dependent steering, such that $w=0$ recovers the ordinary conditional velocity. 

\section{Results}
\label{sec:results}

We train a 617k-parameter \textsc{JetFUEL} model with six message-passing layers and a hidden dimension of 96 on particle clouds from the \jetnet~\cite{kansal2023jetnet} dataset, using the 30 highest-momentum particle constituents in each jet\footnote{To support the generation of variable-sized particle clouds, we use masking within the model, following previous work~\cite{kansalParticleCloudGeneration2022}.}. We train for 994,000 optimizer updates with an effective batch size of 250, using 368,113 gluon, light quark, and top quark jets.

At training time, \jetnet provides the jet attributes in the conditioning vector $c$ for each cloud. At inference time, we sample the jet size from a categorical distribution estimated from the \jetnet training samples and sample the other attributes from a normalizing flow~\cite{papamakarios2021normalizing} fitted before training (see \autoref{appdx:nf} for details). We implement CFG by dropping the conditioning vector $c$ with probability $b=0.1$. At inference, we generate samples with 128 integration steps.

Our preliminary recipe is competitive with MPGAN~\cite{kansalParticleCloudGeneration2022} and several other methods according to the \jetnet metrics reported in \autoref{tab:jetnet30-comparison}.\footnote{We report the classifier-free-guidance (CFG) weight that yields the best FPND value and provide a full sweep in \autoref{sec:addl-results}.}
\begin{table}[H]
    \centering
    \setlength{\abovecaptionskip}{2pt}
    \setlength{\belowcaptionskip}{2pt}
    \caption{Generation metrics on \jetnet-30. The best value for each jet class is bold.}
    \label{tab:jetnet30-comparison}
\begingroup
\scriptsize
\setlength{\tabcolsep}{3pt}
\renewcommand{\arraystretch}{0.70}
\resizebox{0.94\textwidth}{!}{%
\begin{tabular}{@{}clrrrrrr@{}}
\toprule
Jet class & Method & FPND $\downarrow$ & \shortstack{FPD $\downarrow$\\$(\times 10^{-4})$} & \shortstack{W1M $\downarrow$\\$(\times 10^{-4})$} & \shortstack{Mean W1P $\downarrow$\\$(\times 10^{-4})$} & Cov. $\uparrow$ & \shortstack{MMD $\downarrow$\\$(\times 10^{-2})$} \\
\midrule
\multirow{7}{*}{Gluon jets} & MPGAN \cite{kansalParticleCloudGeneration2022} & 0.12 & \textbf{4.1} & 7 & 9 & 0.56 & 3.7 \\
 & FPCD \cite{mikuniFastPointCloud2023} & 0.07 & \textemdash{} & \textbf{3.6} & 3.4 & 0.55 & 3 \\
 & PET generator \cite{bhimji2026foundation} & 0.04 & \textemdash{} & 4.2 & 3.6 & 0.55 & 3 \\
 & OmniLearn \cite{bhimji2026foundation} & \textbf{0.02} & \textemdash{} & 3.8 & \textbf{3.3} & 0.55 & 3 \\
 & PC-JeDi DDIM \cite{leighPCJeDiDiffusionParticle2024} & 0.12 & --- & 9 & 6.6 & 0.54 & 3.64 \\
 & PC-JeDi EM \cite{leighPCJeDiDiffusionParticle2024} & 0.1 & --- & 5.7 & 5.8 & 0.54 & 3.62 \\
\cmidrule(l){2-8}
 & \textsc{JetFUEL} ($w=0$) & 0.0797 & 9.9029 & 8.0979 & 5.0967 & \textbf{0.570} & \textbf{2.6415} \\
\midrule
\multirow{6}{*}{Light quark jets} & MPGAN & 0.35 & 19 & 6 & 49 & 0.5 & 2.6 \\
 & FPCD & 0.08 & \textemdash{} & 5.2 & \textbf{2.7} & 0.49 & 2 \\
 & PET generator & 0.03 & \textemdash{} & 3.9 & 3.5 & 0.54 & 2 \\
 & OmniLearn & \textbf{0.02} & \textemdash{} & \textbf{2.4} & 3.2 & 0.54 & 2 \\
\cmidrule(l){2-8}
 & \textsc{JetFUEL} ($w=0$) & 0.1546 & \textbf{16.0313} & 10.0082 & 10.6199 & \textbf{0.573} & 1.8662 \\
 & \textsc{JetFUEL} ($w=0.25$) & 0.1327 & 39.7412 & 10.5851 & 9.0692 & 0.554 & \textbf{1.8242} \\
\midrule
\multirow{8}{*}{Top jets} & MPGAN & 0.36 & 28 & 6.4 & 21.7 & 0.58 & 7.11 \\
 & FPCD & 0.17 & \textemdash{} & 5.1 & 4.1 & 0.58 & \textbf{5} \\
 & PET generator & 0.07 & \textemdash{} & 4.4 & \textbf{2.9} & 0.58 & \textbf{5} \\
 & OmniLearn & \textbf{0.04} & \textemdash{} & \textbf{4.3} & 3 & 0.58 & \textbf{5} \\
 & PC-JeDi DDIM & 0.28 & --- & 14.8 & 10.1 & \textbf{0.59} & 7.13 \\
 & PC-JeDi EM & 0.15 & --- & 13.6 & 12.1 & \textbf{0.59} & 7.11 \\
\cmidrule(l){2-8}
 & \textsc{JetFUEL} ($w=0$) & 0.3693 & \textbf{8.3087} & 9.9065 & 11.8780 & 0.578 & 5.5700 \\
 & \textsc{JetFUEL} ($w=0.5$) & 0.1744 & 28.2540 & 14.0792 & 13.0853 & 0.589 & 5.5920 \\
\bottomrule
\end{tabular}
}%
\endgroup

\end{table}

\section{Conclusion}
\label{sec:conclusion}

Applying deep learning to scientific domains raises design questions about the extent and manner in which prior knowledge about the domain should be incorporated. In this work, we showed a method that integrates physical knowledge in an end-to-end fashion for generative modeling of jets, where physics guided the model architecture, training objective, choice of data prior, integration paths for generation, and training heuristics such as particle cloud coupling. 

Our results show that this recipe results in stable jet generation, matching foundational work such as MPGAN \cite{kansalParticleCloudGeneration2022} while utilizing only one model trained across multiple jet types. We show that a simple Lorentz-equivariant architecture utilizing only scalar and four-vector features is sufficient for generative modeling, simplifying approaches from prior work \cite{spinner2024lorentz}. We show that despite the degeneracy of the Minkowski metric, exploiting the known scientific structure of jet constituents enables the straightforward application of RFM to jets. The recipe presented in \textsc{JetFUEL} may inform further work on physics-informed generative models for jet physics that leverages the physical interpretability of the recipe to guide further development. Future work can modify and scale the \textsc{JetFUEL} recipe to achieve higher distributional fidelity while continuing to benefit from physical plausibility and interpretability in generative modeling.

\section*{Acknowledgments and Disclosure of Funding}
This work used resources available through the National Research Platform (NRP) at the University of California, San Diego. NRP has been developed, and is supported in part, by funding from the National Science Foundation, from awards 1730158, 1540112, 1541349, 1826967, 2112167, 2100237, and 2120019, as well as additional funding from community partners. The CSUF Titan Supercomputing Center is one of the collaborative partners to contribute to NRP resources. R.K., Z.H., and M.S. are supported by the
California Institute of Technology High Energy Physics under Contract DE-SC0011925 with the
United States Department of Energy, Office of High Energy Physics. A.G. and J.N. are supported by
the DOE Office of Science, Award No. DE-SC0023524, FermiForward Discovery Group, LLC under
Contract No. 89243024CSC000002 with the U.S. Department of Energy, Office of Science, Office of
High Energy Physics, LDRD L2024-066-1, Fermilab, DOE Office of Science, Office of High Energy
Physics “Designing efficient edge AI with physics phenomena” Project (DE-FOA-0002705), DOE
Office of Science, Office of Advanced Scientific Computing Research under the “Real-time Data
Reduction Codesign at the Extreme Edge for Science” Project (DE-FOA-0002501). R.K. and J.N. are
also supported by the AI2050 program at Schmidt Futures (Grant G-23-64934).

\clearpage
\appendix

\bibliographystyle{unsrt}
\bibliography{refs}

@article{buhmannEPiCGANEquivariantPoint2023,
  title = {{{EPiC-GAN}}: {{Equivariant Point Cloud Generation}} for {{Particle Jets}}},
  shorttitle = {{{EPiC-GAN}}},
  author = {Buhmann, Erik and Kasieczka, Gregor and Thaler, Jesse},
  year = 2023,
  month = oct,
  journal = {SciPost Physics},
  volume = {15},
  number = {4},
  eprint = {2301.08128},
  primaryclass = {hep-ph},
  pages = {130},
  issn = {2542-4653},
  doi = {10.21468/SciPostPhys.15.4.130},
  urldate = {2025-03-13},
  archiveprefix = {arXiv}
}

@article{butterMachineLearningLHC2023,
  title = {Machine {{Learning}} and {{LHC Event Generation}}},
  author = {Butter, Anja and Plehn, Tilman and Schumann, Steffen and Badger, Simon and Caron, Sascha and Cranmer, Kyle and Bello, Francesco Armando Di and Dreyer, Etienne and Forte, Stefano and Ganguly, Sanmay and Gon{\c c}alves, Dorival and Gross, Eilam and Heimel, Theo and Heinrich, Gudrun and Heinrich, Lukas and Held, Alexander and H{\"o}che, Stefan and Howard, Jessica N. and Ilten, Philip and Isaacson, Joshua and Jan{\ss}en, Timo and Jones, Stephen and Kado, Marumi and Kagan, Michael and Kasieczka, Gregor and Kling, Felix and Kraml, Sabine and Krause, Claudius and Krauss, Frank and Kr{\"o}ninger, Kevin and Barman, Rahool Kumar and Luchmann, Michel and Magerya, Vitaly and Maitre, Daniel and Malaescu, Bogdan and Maltoni, Fabio and Martini, Till and Mattelaer, Olivier and Nachman, Benjamin and Pitz, Sebastian and Rojo, Juan and Schwartz, Matthew and Shih, David and Siegert, Frank and Stegeman, Roy and Stienen, Bob and Thaler, Jesse and Verheyen, Rob and Whiteson, Daniel and Winterhalder, Ramon and Zupan, Jure},
  year = 2023,
  month = apr,
  journal = {SciPost Physics},
  volume = {14},
  number = {4},
  eprint = {2203.07460},
  primaryclass = {hep-ph},
  pages = {079},
  issn = {2542-4653},
  doi = {10.21468/SciPostPhys.14.4.079},
  urldate = {2025-06-26},
  archiveprefix = {arXiv},
  langid = {english}
}

@misc{FastSimulationATLAS2020,
  title = {Fast Simulation of the {{ATLAS}} Calorimeter System with {{Generative Adversarial Networks}}},
  year = 2020,
  publisher = {CERN},
  address = {Geneva},
  urldate = {2025-03-02}
}

@article{gongEfficientLorentzEquivariant2022,
  title = {An Efficient {{Lorentz}} Equivariant Graph Neural Network for Jet Tagging},
  author = {Gong, Shiqi and Meng, Qi and Zhang, Jue and Qu, Huilin and Li, Congqiao and Qian, Sitian and Du, Weitao and Ma, Zhi-Ming and Liu, Tie-Yan},
  year = 2022,
  month = jul,
  journal = {Journal of High Energy Physics},
  volume = {2022},
  number = {7},
  pages = {30},
  issn = {1029-8479},
  doi = {10.1007/JHEP07(2022)030},
  urldate = {2025-06-30},
  langid = {english}
}

@article{haoLorentzGroupEquivariant2023,
  title = {Lorentz Group Equivariant Autoencoders},
  author = {Hao, Zichun and Kansal, Raghav and Duarte, Javier and Chernyavskaya, Nadezda},
  year = 2023,
  month = jun,
  journal = {The European Physical Journal C},
  volume = {83},
  number = {6},
  eprint = {2212.07347},
  primaryclass = {hep-ex},
  pages = {485},
  issn = {1434-6052},
  doi = {10.1140/epjc/s10052-023-11633-5},
  urldate = {2025-03-13},
  archiveprefix = {arXiv}
}

@misc{hoClassifierFreeDiffusionGuidance2022,
  title = {Classifier-{{Free Diffusion Guidance}}},
  author = {Ho, Jonathan and Salimans, Tim},
  year = 2022,
  month = jul,
  number = {arXiv:2207.12598},
  eprint = {2207.12598},
  primaryclass = {cs},
  publisher = {arXiv},
  doi = {10.48550/arXiv.2207.12598},
  urldate = {2025-08-05},
  archiveprefix = {arXiv}
}

@misc{Introduction,
  title = {Introduction},
  urldate = {2025-03-13},
  howpublished = {https://deepgenerativemodels.github.io/notes/introduction/}
}

@misc{kachAttentionMeanFieldsParticle2023a,
  title = {Attention to {{Mean-Fields}} for {{Particle Cloud Generation}}},
  author = {K{\"a}ch, Benno and {Melzer-Pellmann}, Isabell},
  year = 2023,
  month = may,
  number = {arXiv:2305.15254},
  eprint = {2305.15254},
  primaryclass = {hep-ex},
  publisher = {arXiv},
  doi = {10.48550/arXiv.2305.15254},
  urldate = {2025-03-13},
  archiveprefix = {arXiv}
}

@misc{kachJetFlowGeneratingJets2022,
  title = {{{JetFlow}}: {{Generating Jets}} with {{Conditioned}} and {{Mass Constrained Normalising Flows}}},
  shorttitle = {{{JetFlow}}},
  author = {K{\"a}ch, Benno and Kr{\"u}cker, Dirk and {Melzer-Pellmann}, Isabell and Scham, Moritz and Schnake, Simon and {Verney-Provatas}, Alexi},
  year = 2022,
  month = nov,
  number = {arXiv:2211.13630},
  eprint = {2211.13630},
  primaryclass = {hep-ex},
  publisher = {arXiv},
  doi = {10.48550/arXiv.2211.13630},
  urldate = {2025-03-13},
  archiveprefix = {arXiv}
}

@article{kansalParticleCloudGeneration2022,
  title={Particle cloud generation with message passing generative adversarial networks},
  author={Kansal, Raghav and Duarte, Javier and Su, Hao and Orzari, Breno and Tomei, Thiago and Pierini, Maurizio and Touranakou, Mary and Gunopulos, Dimitrios and others},
  journal={Advances in Neural Information Processing Systems},
  volume={34},
  pages={23858--23871},
  year={2021}
}

@article{leighPCJeDiDiffusionParticle2024,
  title = {{{PC-JeDi}}: {{Diffusion}} for {{Particle Cloud Generation}} in {{High Energy Physics}}},
  shorttitle = {{{PC-JeDi}}},
  author = {Leigh, Matthew and Sengupta, Debajyoti and Qu{\'e}tant, Guillaume and Raine, John Andrew and Zoch, Knut and Golling, Tobias},
  year = 2024,
  month = jan,
  journal = {SciPost Physics},
  volume = {16},
  number = {1},
  eprint = {2303.05376},
  primaryclass = {hep-ph},
  pages = {018},
  issn = {2542-4653},
  doi = {10.21468/SciPostPhys.16.1.018},
  urldate = {2025-06-23},
  archiveprefix = {arXiv}
}

@article{liDoesLorentzsymmetricDesign2024,
  title = {Does {{Lorentz-symmetric}} Design Boost Network Performance in Jet Physics?},
  author = {Li, Congqiao and Qu, Huilin and Qian, Sitian and Meng, Qi and Gong, Shiqi and Zhang, Jue and Liu, Tie-Yan and Li, Qiang},
  year = 2024,
  month = mar,
  journal = {Physical Review D},
  volume = {109},
  number = {5},
  pages = {056003},
  issn = {2470-0010, 2470-0029},
  doi = {10.1103/PhysRevD.109.056003},
  urldate = {2025-05-20},
  langid = {english}
}

@inproceedings{lipmanFlowMatchingGenerative2023,
  title={Flow matching for generative modeling},
  author={Lipman, Yaron and Chen, Ricky T. Q. and Ben-Hamu, Heli and Nickel, Maximilian and Le, Matt},
  booktitle={International Conference on Learning Representations},
  year={2023}
}

@article{mikuniFastPointCloud2023,
  title = {Fast {{Point Cloud Generation}} with {{Diffusion Models}} in {{High Energy Physics}}},
  author = {Mikuni, Vinicius and Nachman, Benjamin and Pettee, Mariel},
  year = 2023,
  month = aug,
  journal = {Physical Review D},
  volume = {108},
  number = {3},
  eprint = {2304.01266},
  primaryclass = {hep-ph},
  pages = {036025},
  issn = {2470-0010, 2470-0029},
  doi = {10.1103/PhysRevD.108.036025},
  urldate = {2025-03-13},
  archiveprefix = {arXiv}
}

@article{oliveiraLearningParticlePhysics2017,
  title = {Learning {{Particle Physics}} by {{Example}}: {{Location-Aware Generative Adversarial Networks}} for {{Physics Synthesis}}},
  shorttitle = {Learning {{Particle Physics}} by {{Example}}},
  author = {de Oliveira, Luke and Paganini, Michela and Nachman, Benjamin},
  year = 2017,
  month = nov,
  journal = {Computing and Software for Big Science},
  volume = {1},
  number = {1},
  eprint = {1701.05927},
  primaryclass = {stat},
  pages = {4},
  issn = {2510-2036, 2510-2044},
  doi = {10.1007/s41781-017-0004-6},
  urldate = {2025-02-21},
  archiveprefix = {arXiv}
}

@article{touranakouParticlebasedFastJet2022,
  title = {Particle-Based {{Fast Jet Simulation}} at the {{LHC}} with {{Variational Autoencoders}}},
  author = {Touranakou, Mary and Chernyavskaya, Nadezda and Duarte, Javier and Gunopulos, Dimitrios and Kansal, Raghav and Orzari, Breno and Pierini, Maurizio and Tomei, Thiago and Vlimant, Jean-Roch},
  year = 2022,
  month = sep,
  journal = {Machine Learning: Science and Technology},
  volume = {3},
  number = {3},
  eprint = {2203.00520},
  primaryclass = {physics},
  pages = {035003},
  issn = {2632-2153},
  doi = {10.1088/2632-2153/ac7c56},
  urldate = {2025-08-08},
  archiveprefix = {arXiv}
}

@article{karagiorgi2022machine,
  title={Machine learning in the search for new fundamental physics},
  author={Karagiorgi, Georgia and Kasieczka, Gregor and Kravitz, Scott and Nachman, Benjamin and Shih, David},
  journal={Nature Reviews Physics},
  volume={4},
  number={6},
  pages={399--412},
  year={2022},
  publisher={Nature Publishing Group UK London}
}

@article{apostolakis2022detector,
  title={Detector simulation challenges for future accelerator experiments},
  author={Apostolakis, John and Bandieramonte, Marilena and Banerjee, Sunanda and Bartosik, Nazar and Corti, Gloria and Cosmo, Gabriele and Elvira, V Daniel and Evans, Thomas and Gheata, Andrei and Pagan Griso, Simone and others},
  journal={Frontiers in Physics},
  volume={10},
  pages={913510},
  year={2022},
  publisher={Frontiers Media SA}
}

@article{guest2018deep,
  title={Deep learning and its application to {LHC} physics},
  author={Guest, Dan and Cranmer, Kyle and Whiteson, Daniel},
  journal={Annual Review of Nuclear and Particle Science},
  volume={68},
  number={1},
  pages={161--181},
  year={2018},
  publisher={Annual Reviews}
}

@article{jumper2021highly,
  title={Highly accurate protein structure prediction with {AlphaFold}},
  author={Jumper, John and Evans, Richard and Pritzel, Alexander and Green, Tim and Figurnov, Michael and Ronneberger, Olaf and Tunyasuvunakool, Kathryn and Bates, Russ and {\v{Z}}{\'\i}dek, Augustin and Potapenko, Anna and others},
  journal={Nature},
  volume={596},
  number={7873},
  pages={583--589},
  year={2021},
  publisher={Nature Publishing Group UK London}
}

@article{kansal2023jetnet,
  title={{JetNet}: A {Python} package for accessing open datasets and benchmarking machine learning methods in high energy physics},
  author={Kansal, Raghav and Pareja, Carlos and Hao, Zichun and Duarte, Javier},
  journal={Journal of Open Source Software},
  volume={8},
  number={90},
  pages={5789},
  year={2023}
}

@article{larkoski2020jet,
  title={Jet substructure at the Large Hadron Collider: a review of recent advances in theory and machine learning},
  author={Larkoski, Andrew J and Moult, Ian and Nachman, Benjamin},
  journal={Physics Reports},
  volume={841},
  pages={1--63},
  year={2020},
  publisher={Elsevier}
}

@article{bronstein2021geometric,
  title={Geometric deep learning: Grids, groups, graphs, geodesics, and gauges},
  author={Bronstein, Michael M and Bruna, Joan and Cohen, Taco and Veli{\v{c}}kovi{\'c}, Petar},
  journal={arXiv preprint arXiv:2104.13478},
  year={2021}
}

@article{papamakarios2021normalizing,
  title={Normalizing flows for probabilistic modeling and inference},
  author={Papamakarios, George and Nalisnick, Eric and Rezende, Danilo Jimenez and Mohamed, Shakir and Lakshminarayanan, Balaji},
  journal={Journal of Machine Learning Research},
  volume={22},
  number={57},
  pages={1--64},
  year={2021}
}

@article{lou2020neural,
  title={Neural manifold ordinary differential equations},
  author={Lou, Aaron and Lim, Derek and Katsman, Isay and Huang, Leo and Jiang, Qingxuan and Lim, Ser Nam and De Sa, Christopher M},
  journal={Advances in Neural Information Processing Systems},
  volume={33},
  pages={17548--17558},
  year={2020}
}

@inproceedings{murnane2026equivariant,
  title={Equivariant graph neural networks for charged particle tracking},
  author={Murnane, Daniel and Thais, Savannah and Thete, Ameya},
  booktitle={Journal of Physics: Conference Series},
  volume={3206},
  number={1},
  pages={012093},
  year={2026},
  organization={IOP Publishing}
}

@article{durkan2019neural,
  title={Neural spline flows},
  author={Durkan, Conor and Bekasov, Artur and Murray, Iain and Papamakarios, George},
  journal={Advances in neural information processing systems},
  volume={32},
  year={2019}
}

@article{spinner2026lorentz,
  title={Lorentz local canonicalization: How to make any network {Lorentz}-equivariant},
  author={Spinner, Jonas and Favaro, Luigi and Lippmann, Peter and Pitz, Sebastian and Gerhartz, Gerrit and Plehn, Tilman and Hamprecht, Fred},
  journal={Advances in Neural Information Processing Systems},
  volume={38},
  pages={28779--28807},
  year={2026}
}

@inproceedings{peebles2023scalable,
  title={Scalable diffusion models with transformers},
  author={Peebles, William and Xie, Saining},
  booktitle={2023 IEEE/CVF International Conference on Computer Vision (ICCV)},
  pages={4172--4182},
  year={2023},
  organization={IEEE}
}

@inproceedings{chen2024flow,
  title={Flow matching on general geometries},
  author={Chen, Ricky TQ and Lipman, Yaron},
  booktitle={International Conference on Learning Representations},
  volume={2024},
  pages={47922--47945},
  year={2024}
}

@article{bhimji2026foundation,
  title={Foundation model framework for all tasks involving jet physics},
  author={Bhimji, Wahid and Harris, Chris and Mikuni, Vinicius and Nachman, Benjamin},
  journal={Physical Review D},
  volume={113},
  number={3},
  pages={032020},
  year={2026},
  publisher={APS}
}

@article{spinner2024lorentz,
  title={{Lorentz}-equivariant geometric algebra transformers for high-energy physics},
  author={Spinner, Jonas and Bres{\'o}, Victor and De Haan, Pim and Plehn, Tilman and Thaler, Jesse and Brehmer, Johann},
  journal={Advances in neural information processing systems},
  volume={37},
  pages={22178--22205},
  year={2024}
}

@article{klein2023equivariant,
  title={Equivariant flow matching},
  author={Klein, Leon and Kr{\"a}mer, Andreas and No{\'e}, Frank},
  journal={Advances in Neural Information Processing Systems},
  volume={36},
  pages={59886--59910},
  year={2023}
}

\newpage

\section{Riemannian Flow Matching}\label{appendix:rfm}

\begin{theorem}
\label{thm:positive-energy-light-cone-degenerate}
The Minkowski metric restricted to the positive-energy light cone $C^+ =\left\{p\in\mathbb R^{1,3}: \langle p,p\rangle_M=0,\ p^0>0 \right\}$ is not positive definite.
\end{theorem}
\begin{proof}
Let $p\in C^+$ and $\lambda>0$ with $\lambda\neq1$. The points $p$ and $\lambda p$ are distinct elements of $C^+$, and their nonzero displacement $(\lambda-1)p$ is tangent to the cone along the ray through $p$. However, the norm induced by the Minkowski metric is zero:
\begin{align*}
\|\lambda p-p\|_M^2 &= \|(\lambda-1)p\|_M^2\\
&= (\lambda-1)^2\|p\|_M^2\\
&= 0.
\end{align*}
Hence the induced Minkowski metric assigns zero norm to a nonzero tangent direction. It is therefore not positive definite and cannot support Riemannian flow matching.
\end{proof} 

\subsection{Radius-\texorpdfstring{$m$}{m} hyperboloid geometry}
\label{appendix:hyperboloid-geometry}

For a regulator mass $m>0$, recall the forward hyperboloid
\begin{equation}
    \mathcal H_m
    =
    \left\{p\in\mathbb R^{1,3}:\langle p,p\rangle_M=m^2,\ p^0>0\right\}.
\end{equation}
This is the radius-$m$ version of the unit hyperboloid. In particular, setting
$x=p/m$ gives $\langle x,x\rangle_M=1$. Thus, the standard unit-hyperboloid
expressions in \cite[Table~2]{lou2020neural} extend to $\mathcal H_m$ by this
rescaling, with distances restored in units of $m$.

We now make the rescaling explicit. Let $p,q\in\mathcal H_m$ and define
\begin{equation}
    x=\frac{p}{m},
    \qquad
    y=\frac{q}{m},
    ,\label{eq:unit-hyperboloid-rescaling}
\end{equation}
so that $x,y\in\mathcal H_1$. Differentiating the constraint
$\langle p,p\rangle_M=m^2$ shows that
\begin{equation}
    T_p\mathcal H_m
    =
    \left\{u\in\mathbb R^{1,3}:\langle p,u\rangle_M=0\right\}.
\end{equation}
We equip this tangent space with the positive-definite metric
\begin{equation}
    g_p(u,w)=-\langle u,w\rangle_M,
    \qquad
    \lVert u\rVert_p=\sqrt{-\langle u,u\rangle_M}.
    \label{eq:hyperboloid-metric}
\end{equation}
Under the same rescaling, $u\in T_p\mathcal H_m$ corresponds to
$v=u/m\in T_x\mathcal H_1$, because
$\langle x,v\rangle_M=\langle p,u\rangle_M/m^2=0$. Moreover,
\begin{equation}
    \lVert v\rVert_x
    =
    \sqrt{-\left\langle\frac{u}{m},\frac{u}{m}\right\rangle_M}
    =
    \frac{\lVert u\rVert_p}{m}.
    \label{eq:tangent-norm-rescaling}
\end{equation}

After translating the sign convention of \cite[Table~2]{lou2020neural} to our
$(+,-,-,-)$ convention, define
\begin{equation}
    \gamma
    =
    \langle x,y\rangle_M
    =
    \frac{\langle p,q\rangle_M}{m^2},
    \qquad
    \alpha=\operatorname{arcosh}(\gamma).
    \label{eq:hyperboloid-gamma}
\end{equation}
The corresponding unit-hyperboloid formulas are
\begin{align}
    d_1(x,y)
    &=\alpha,
    \label{eq:unit-hyperboloid-distance}\\
    \Log_x^{(1)}(y)
    &=
    \frac{\alpha}{\sinh(\alpha)}
    \left(y-\gamma x\right),
    \label{eq:unit-hyperboloid-log-map}\\
    \Exp_x^{(1)}(v)
    &=
    \cosh\!\left(\lVert v\rVert_x\right)x
    +
    \sinh\!\left(\lVert v\rVert_x\right)
    \frac{v}{\lVert v\rVert_x}.
    \label{eq:unit-hyperboloid-exp-map}
\end{align}

The scaling map $F_m(x)=mx$ multiplies every curve length by $m$. Therefore,
the radius-$m$ geodesic distance is
\begin{align}
    d_g(p,q)
    &=m\,d_1\!\left(\frac{p}{m},\frac{q}{m}\right)\\
    &=m\alpha\\
    &=m\,\operatorname{arcosh}\!\left(
        \frac{\langle p,q\rangle_M}{m^2}
      \right).
    \label{eq:hyperboloid-distance}
\end{align}
The logarithmic map is a tangent displacement and scales by the same factor.
Substituting $x=p/m$ and $y=q/m$ into the unit-hyperboloid map yields
\begin{align}
    \Log_p(q)
    &=m\Log_x^{(1)}(y)\\
    &=m\frac{\alpha}{\sinh(\alpha)}
      \left(\frac{q}{m}-\gamma\frac{p}{m}\right)\\
    &=
    \frac{\alpha}{\sinh(\alpha)}
    \left(q-\gamma p\right),
    \qquad
    \gamma=\frac{\langle p,q\rangle_M}{m^2}.
    \label{eq:hyperboloid-log-map}
\end{align}
Thus $\Log_p(q)\in T_p\mathcal H_m$ and
$\lVert\Log_p(q)\rVert_p=d_g(p,q)$.

Finally, we obtain the radius-$m$ exponential map by mapping $p$ and $u$ to
$x=p/m$ and $v=u/m$, applying the unit-hyperboloid exponential map, and scaling
the result back by $m$:
\begin{align}
    \Exp_p(u)
    &=m\Exp_x^{(1)}\!\left(\frac{u}{m}\right)\\
    &=m\left[
       \cosh\!\left(\frac{\lVert u\rVert_p}{m}\right)\frac{p}{m}
       +
       \sinh\!\left(\frac{\lVert u\rVert_p}{m}\right)
       \frac{u/m}{\lVert u\rVert_p/m}
       \right]\\
    &=
    \cosh\!\left(\frac{\lVert u\rVert_p}{m}\right)p
    +
    m\sinh\!\left(\frac{\lVert u\rVert_p}{m}\right)
    \frac{u}{\lVert u\rVert_p}.
    \label{eq:hyperboloid-exp-map}
\end{align}

\subsection{Tangent projection}
\label{appendix:tangent-projection}

An arbitrary ambient four-vector $v\in\mathbb R^{1,3}$ can be projected onto
$T_p\mathcal H_m$ using the Minkowski-orthogonal projection
\begin{equation}
    \Pi_p(v)
    =
    v-\frac{\langle p,v\rangle_M}{m^2}p.
    \label{eq:hyperboloid-tangent-projection}
\end{equation}

\subsection{Numerical implementation}

We evaluate the mass-shell projection, geodesic distance, logarithmic and
exponential maps, and RFM loss in double precision. We protect the domain of
$\operatorname{arcosh}$ and divisions by vanishing tangent norms with small
numerical floors. Likewise, the denominator $1-t$ in the conditional target is
bounded away from zero, and integration ends
at $t=0.99999$ rather than evaluating the expression at $t=1$. Finally, after
each exponential-Euler update, we recompute the energy as
$p^0=\sqrt{\lVert\mathbf p\rVert_2^2+m^2}$ to remove accumulated floating-point
drift from $\mathcal H_m$.

In our experiments, we set the regulator mass to $m=0.1$ in
model-scaled units. As $m\to 0$, the hyperboloid approaches the light
cone, but we face increasing numerical instability, since tangent projection
and the logarithm map involve factors of $1/m^2$, and the exponential
map depends on $\lVert u\rVert_p/m$. We found $m=0.1$ to achieve a satisfactory balance between proximity to the light cone and numerical stability during training.

\section{Network architecture}
\label{appdx:network-architecture}

\subsection{Algorithm description}
\label{appdx:algorithm-description}
\begingroup
\setlength{\intextsep}{2pt}
\setlength{\abovecaptionskip}{1pt}
\setlength{\belowcaptionskip}{1pt}

Below, we present the \textsc{JetFUEL} model architecture discussed in \autoref{sec:method} in algorithm form. Recall that $t$ is the flow matching integration time, $c$ is the conditioning vector, $r_0$ and $r_1$ are reference vectors corresponding to $(1, 0, 0, 0)$ and the jet axis four-momentum, respectively. $\phi$ always corresponds to a multilayer perceptron. Before Minkowski invariants are passed to an MLP, we apply the signed logarithm
\[
\psi(z)=\operatorname{sgn}(z)\log(1+|z|)
\]
elementwise, which compresses their large dynamic range while preserving sign.

For completeness, we denote by $\mathcal{N}(i)$ the neighborhood of each constituent, and in practice utilize a fully-connected approach where each particle is connected to all other constituents in the cloud, which has a maximum size of thirty in the dataset.

In the algorithm, lines 1--2 initialize directions $y$ and hidden scalar values $h$ for each constituent. Lines 3--11 consist of the LorentzNet-style \cite{gongEfficientLorentzEquivariant2022} message-passing phase, where we calculate messages between each pair of particles using Lorentz-invariant features and update particle directions as a linear combination of latent particle displacements $y_i^\ell - y_j^\ell$, weighted by a scalar value $\epsilon_y = 10^{-2}$. In lines 12--16, we perform a final particle-level readout, calculating a final per-particle change $v_i^\text{part}$ as a sum of on-shell directions $q_{ij}$ weighted by coefficients $a_{ij}$, with these coefficients also being calculated from Lorentz-invariant features. In lines 17--19, we calculate a weighted update along the directions supplied by $r_0$ and $r_1$. In line 20, we combine the directions supplied by other particles with the reference directions for the final per-particle updates. 

\begin{algorithm}[H]

\caption{\textsc{JetFUEL} vector field $V_\theta(X,t,N,c,r_0,r_1)$}
\label{alg:network}
\tiny
\begin{algorithmic}[1]
\Require $X=(x_1,\ldots,x_N)$, $t$, $c$, $(r_0,r_1)$
\Ensure $V_\theta(X,t,N,c,r_0,r_1)\in T_X\mathcal{H}_m^N$
\State $y_i^0\gets x_i$ and $\tau(t)\gets[t,\sin(\pi t),\cos(\pi t)]$ for all $i$
\State $h_i^0\gets\phi_{\rm seed}([\psi(\langle y_i^0,r_0\rangle_M),\psi(\langle y_i^0,r_1\rangle_M)])+\phi_c(c)-\phi_t(\tau(t))$
\For{$\ell=0,\ldots,L-1$}
  \State $\bar h_i^\ell\gets\operatorname{LayerNorm}(h_i^\ell)$
  \For{each $j\in\mathcal{N}(i)$}
    \State $\xi_{ij}^\ell\gets[\psi(\|y_i^\ell-y_j^\ell\|_M^2),\psi(\langle y_i^\ell,y_j^\ell\rangle_M)]$
    \State $M_{ij}^\ell\gets\phi_e^\ell(\bar h_i^\ell,\bar h_j^\ell,\xi_{ij}^\ell)$; \quad $\alpha_{ij}^\ell\gets\phi_x^\ell(M_{ij}^\ell)$
  \EndFor
  \State $h_i^{\ell+1}\gets h_i^\ell+\phi_h^\ell\!\left(\bar h_i^\ell,|\mathcal{N}(i)|^{-1/2}\sum_{j\in\mathcal{N}(i)}M_{ij}^\ell\right)$
  \State $y_i^{\ell+1}\gets y_i^\ell+\varepsilon_y|\mathcal{N}(i)|^{-1/2}\sum_{j\in\mathcal{N}(i)}\alpha_{ij}^\ell(y_i^\ell-y_j^\ell)$
\EndFor
\For{each $j\in\mathcal{N}(i)$}
  \State $\xi_{ij}^L\gets[\psi(\|y_i^L-y_j^L\|_M^2),\psi(\langle y_i^L,y_j^L\rangle_M)]$
  \State $a_{ij}\gets\phi_f(h_i^L,h_j^L,\xi_{ij}^L)$; \quad $q_{ij}\gets\Log_{x_i}(x_j)/(m+\dist(x_i,x_j))$
\EndFor
\State $v_i^{\rm part}\gets|\mathcal{N}(i)|^{-1/2}\sum_{j\in\mathcal{N}(i)}a_{ij}q_{ij}$
\State $(b_{i0},b_{i1})\gets\phi_r(h_i^L,[\psi(\langle y_i^L,r_0\rangle_M),\psi(\langle y_i^L,r_1\rangle_M)])$
\State $q_{ir}^{\rm ref}\gets\Proj_{x_i}(r_r)/(m+\|\Proj_{x_i}(r_r)\|_{x_i})$, for $r\in\{0,1\}$
\State $v_i^{\rm ref}\gets b_{i0}q_{i0}^{\rm ref}+b_{i1}q_{i1}^{\rm ref}$
\State $v_{\theta,i}\gets v_i^{\rm part}+v_i^{\rm ref}\in T_{x_i}\HH_m$
\State \Return $V_\theta=(v_{\theta,1},\ldots,v_{\theta,N})$
\end{algorithmic}
\end{algorithm}

\subsubsection{\texorpdfstring{Multilayer perceptron $\phi$ shapes}{phi shapes}}

\scriptsize
\setlength{\abovedisplayskip}{2pt}
\setlength{\belowdisplayskip}{2pt}
In our experiments, $d_h=96$, $\sigma=\operatorname{SiLU}$, and $d_c$ is the conditioning
dimension.

\begin{align}
 \phi_c &: d_c\to d_h\to d_h,&
 \phi_t &: 3\to d_h\to d_h,&
 \phi_{\rm seed}&:2\to d_h\to d_h,\\
 \phi_e^\ell &: (2d_h+2)\to d_h\to d_h,&
 \phi_h^\ell &:2d_h\to2d_h\to d_h,\\
 \phi_x^\ell &:d_h\to d_h\to1,&
 \phi_f &: (2d_h+2)\to d_h\to1,&
 \phi_r &: (d_h+2)\to d_h\to2.
 \label{eq:mlp-widths}
\end{align}

\begin{equation}
 \varepsilon_y=10^{-2},\qquad
 W_{\mathrm{out}}^{\phi_x,\phi_f,\phi_r}\sim\mathcal{N}(0,10^{-6}),
 \qquad b_{\mathrm{out}}^{\phi_x,\phi_f,\phi_r}=0.
\end{equation}
\endgroup

\subsection{Network equivariance}
\label{sec:appdx-equivariance}

\begin{figure}[htbp]
    \centering
    \begin{minipage}{0.49\textwidth}
        \centering
        \includegraphics[width=\linewidth]{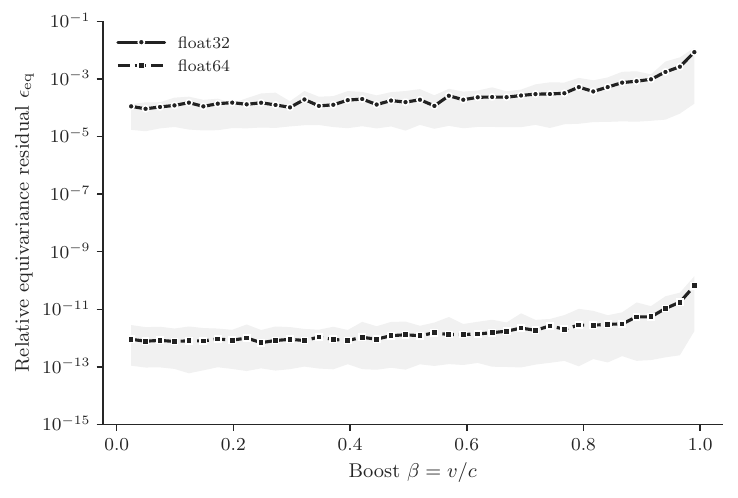}
    \end{minipage}\hfill
    \begin{minipage}{0.49\textwidth}
        \centering
        \includegraphics[width=\linewidth]{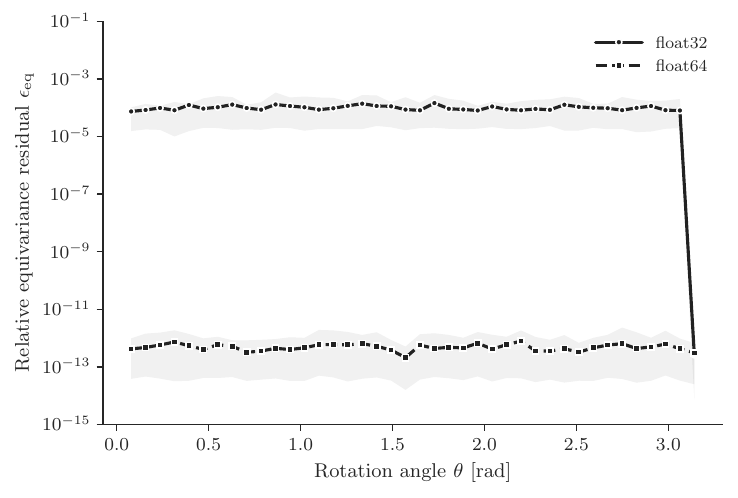}
    \end{minipage}
    \caption{Equivariance error of \textsc{JetFUEL} under Lorentz boosts (left) and
spatial rotations (right), evaluated on class-balanced, held-out \jetnet
constituent clouds projected onto the regulator mass shell. The errors are negligible
and somewhat sensitive to numerical precision.}
    \label{fig:boost-rotation-equivariance}
\end{figure}
For each jet and each group action $G$ (either a Lorentz boost or a spatial rotation), we calculate the relative equivariance error $\epsilon_{\mathrm{eq}}(G)=\left\|M\odot\left[V_\theta(Gx,t,N,c,Gr_0,Gr_1)-G V_\theta(x,t,N,c,r_0,r_1)\right]\right\|_2/\left\|M\odot G V_\theta(x,t,N,c,r_0,r_1)\right\|_2$, where $M$ excludes masked particles, and present the results in \autoref{fig:boost-rotation-equivariance}. We use
$G=\Lambda_x(\beta)$ for boosts and $G=R_y(\omega)$ for rotations, and
report the arithmetic mean over 48 jets, with 16 jets from each of the
gluon, light quark, and top quark classes. Both reference vectors are
transformed jointly with the constituent cloud; for spatial rotations,
$Rr_0=r_0$ because $r_0=(1,0,0,0)$.

\subsection{Rotation-equivariant flows from a rotation-invariant prior generate rotation-invariant distributions}
\label{appdx:equivariant-flow-invariance}
\begin{theorem}
\label{thm:equivariant-flow-preserves-invariance}
Let \(G=\mathrm{SO}(3)\) act on the sample space, let \(p_0\) be a
\(G\)-invariant prior, and let \(\Phi_t\) be a \(G\)-equivariant flow map:
\[
R_\#p_0=p_0,
\qquad
\Phi_t(Rx)=R\Phi_t(x)
\quad
\text{for every }R\in G.
\]
Then the generated distribution \(p_t=(\Phi_t)_\#p_0\) is also
\(G\)-invariant.
\end{theorem}

\begin{proof}
Here \(f_\#p\) denotes the pushforward of \(p\) through \(f\), i.e., the
distribution of \(f(X)\) when \(X\sim p\). Equivariance implies
\(R\circ\Phi_t=\Phi_t\circ R\). Therefore,
\[
R_\#p_t
=
R_\#(\Phi_t)_\#p_0
=
(R\circ\Phi_t)_\#p_0
=
(\Phi_t\circ R)_\#p_0
=
(\Phi_t)_\#R_\#p_0
=
(\Phi_t)_\#p_0
=
p_t.
\]
Thus, \(p_t\) is rotation-invariant.
\end{proof}

\subsection{Scaling and numerical stability}

We divide all four-momentum inputs to the network by a constant scale factor, which preserves Lorentz equivariance while enabling stable training. In practice, we set this value to the mean of the four componentwise standard deviations of real training constituents' \((E,p_x,p_y,p_z)\), which comes out to 58.724~GeV in our data.

\section{Physical prior}
\label{sec:physical-prior}

In Section~\ref{sec:jet-direction}, we discussed the choice of prior for the directions \(\eta_i\) and \(\phi_i\), motivating sampling these values for each constituent from Gaussians around the jet axis with width 0.15. We additionally follow a physics-informed approach to sample the other constituent characteristics. 

\begin{enumerate}
    \item For each constituent, sample \(\eta_i\) and \(\phi_i\) around the jet axis with Gaussian width \(0.15\).
    \item For each constituent, sample \(z_i\sim\mathcal N(0,1)\). Then make these values positive through exponentiation and normalize as \(s_i=e^{z_i}/\sum_j e^{z_j}\).
    \item Set \begin{align*}
        p_{T,i}&=\frac{s_i\,p_{T,J}}
{\left|\sum_j s_j(\cos\phi_j,\sin\phi_j)\right|}\\
E_i&=p_{T,i}\cosh\eta_i
    \end{align*}
This denominator makes the vector sum of constituent transverse momenta have the sampled jet \(p_{T,J}\).
\end{enumerate}

This gives a jet-axis-aligned, physically motivated prior. After rescaling, we project it onto the regulator mass shell used for flow matching.

\section{Normalizing flow for jet attributes}
\label{appdx:nf}

While \jetnet provides global jet attributes with each target point cloud during training, we must sample these attributes at inference time to generate new particle clouds.

We sample the multiplicity of particles in a jet from a simple categorical distribution calculated empirically from jets seen during training, conditioned on the jet type.

Conditioned on the jet type and normalized multiplicity, we learn a conditional normalizing flow \cite{papamakarios2021normalizing} that models the distribution of $\left(\eta_J,\,
        \log p_{T,J},\,
        \log\frac{m_J}{p_{T,J}}\right)$.  We use a three-dimensional standard Gaussian base distribution and ten autoregressive rational-quadratic spline transformations \cite{papamakarios2021normalizing, durkan2019neural}, interleaved with learned linear permutations. Each spline conditioner contains four hidden layers of width \(128\). The model is trained by maximum likelihood on the \jetnet training split, with early stopping based on validation likelihood. At inference, we sample the jet azimuth independently as
\(\phi_J\sim\mathcal U(0,2\pi)\). 

The resulting attributes determine the constituent mask, conditioning vector \(c\), axis-aligned prior, and jet reference four-vector used by the constituent generator.

\section{Additional results}
\label{sec:addl-results}

\subsection{Distributional fidelity}

\begin{figure}[H]
    \centering
    \input{sections/result_distributions}
    \caption{Generated and held-out \jetnet distributions for gluon, light quark, and top quark jets.}
    \label{fig:addl-distribution-fidelity}
\end{figure}

\subsection{Inference-time scaling}

\autoref{tab:addl-inference-scaling} summarizes how FPND changes with the number of integration steps and the classifier-free guidance weight. The strong returns to scaling integration steps from 32 to 64 suggest that the learned vector field is highly nontrivial and benefits from fine-grained integration, and we report 128 integration steps in our headline results. As for the classifier-free guidance weight, higher weights are required for light quark and top quark jets. One potential reason for this may be that the multiplicity of constituents in the unconditional training distribution skews heavily towards 30, and the multiplicity of gluon jets tends to be close to or exactly 30, such that less guidance is required to enforce adherence to the resulting jet structure. Light quark and top quark jets, however, may more frequently require stronger steering to faithfully reproduce the sparser jet structure.

\begin{table}[H]
    \centering
\begingroup
\scriptsize
\setlength{\tabcolsep}{1.5pt}
\renewcommand{\arraystretch}{0.82}
\noindent\makebox[\textwidth][c]{%
\begin{minipage}[t]{0.38\textwidth}
\centering
\begin{tabular}{rrrrr}
\toprule
Steps & $g$ & $q$ & $t$ & Time (s) \\
\midrule
32  & 0.4059 & 0.6132 & 0.9646 &  992.1 \\
64  & 0.1420 & 0.2679 & 0.5244 & 1853.8 \\
128 & \textbf{0.0797} & \textbf{0.1546} & \textbf{0.3693} & 3713.5 \\
\bottomrule
\end{tabular}
\end{minipage}%
\hspace{1.25em}%
\begin{minipage}[t]{0.25\textwidth}
\centering
\begin{tabular}{rrrr}
\toprule
$w$ & $g$ & $q$ & $t$ \\
\midrule
0.00 & \textbf{0.1420} & 0.2679 & 0.5244 \\
0.25 & 0.3182 & \textbf{0.2137} & 0.2801 \\
0.50 & 0.5255 & 0.3138 & \textbf{0.2019} \\
0.75 & 0.7063 & 0.5796 & 0.2582 \\
1.00 & 0.9160 & 1.6340 & 0.4261 \\
2.00 & 4.5388 & 5.7217 & 2.1008 \\
3.00 & 25.0270 & 15.1774 & 8.4812 \\
\bottomrule
\end{tabular}
\end{minipage}%
}
\endgroup

    \caption{FPND as a function of integration steps (left) and classifier-free guidance weight $w$ (right). $g$, $q$, and $t$ correspond to gluon jets, light quark jets, and top quark jets respectively. While we always randomly drop out the conditioning vector during training to enable CFG at inference time, the results in the first table are reported based on jets generated without utilizing CFG. The results in the second table, from varying the CFG weight, correspond to inference-time integration using only 64 steps. Timing numbers are reported based on one A40 GPU with 48 GB of memory and a fixed batch size across integration step ablation experiments.}
    \label{tab:addl-inference-scaling}
\end{table}

\end{document}